\documentclass[11pt]{article}

\usepackage{acl-ijcnlp2021-templates/acl2021}

\aclfinalcopy
\usepackage{times}
\usepackage{latexsym}

\usepackage[T1]{fontenc}
\usepackage{tabularx}
\usepackage{graphicx}

\title{Generating Summaries for Scientific Paper Review}
\author{Ana-Sabina Uban \\
  Faculty of Mathematics and Computer Science, \\ University of Bucharest \\
  \texttt{auban@fmi.unibuc.ro} \\\And
  Cornelia Caragea \\
  Computer Science, \\ University of Illinois at Chicago \\
  \texttt{cornelia@uic.edu} \\}

\date{}

\begin{document}
\maketitle


\begin{abstract}
The review process is essential to ensure the quality of publications. 
Recently, the increase of submissions for top venues in machine learning and NLP has caused a problem of excessive burden on reviewers and has often caused concerns regarding how this may not only overload reviewers, but also may affect the quality of the reviews.
An automatic system for assisting with the reviewing process 
could be a solution for ameliorating the problem. In this paper, we explore automatic review summary generation for scientific papers. We posit that neural language models have the potential to be valuable candidates for this task. In order to test this hypothesis, we release a new dataset of scientific papers and their reviews, collected from papers published in the NeurIPS conference from 2013 to 2020. We evaluate state of the art neural summarization models, present initial results on the feasibility of automatic review summary generation, and propose directions for the future.
\end{abstract}

\section{Introduction}

Reviewing is at the center of the scientific publication process, and the quality of publications is dependent on it. In many scientific fields, including natural language processing and machine learning, submissions for publication are reviewed using a peer review system. Recently, these fields are seeing increasing volumes of submissions each year, especially in high reputation venues. This has created an issue of over-burdening of reviewers, which is not only a problem for the quality of life of scientists, but also consequently affects the quality of the reviews. 
With ever increasing volume of new results in these fields, submissions for publication are expected to multiply still, and the problem is only expected to deepen, which is raising concerns in the scientific community \cite{rogers2020can}.

One avenue for ameliorating this problem is relying on artificial intelligence to assist with the process, in order to remove some of the burden from the human reviewers. A possibility would be to generate reviews or article summaries automatically, in order to speed up the human's understanding of the paper, or to assist with parts of the review writing, e.g., a few sentences summary. 

Text generation has seen impressive improvements in recent years, being one of the most active fields in NLP, with the highest leaps in performance of newly published models. Models such as BERT \cite{devlin2019bert}, GPT-3 \cite{brown2020language} have shown impressive results for text generation, as well as for other tasks, acting as language models which can generalize for a wide range of tasks in NLP with only little fine-tuning. 

Text summarization is a problem of text generation. Depending on the approach, summarization can be extractive \cite{zheng-lapata-2019-sentence} or abstractive \cite{see-etal-2017-get,nallapati-etal-2016-abstractive}. Extractive summarization is performed by selecting key sentences from the original text, while abstractive summarization tackles the more difficult problem of generating novel text that summarizes a given input---the problem we are interested in and explore in this paper. 
As for text generation in general, state-of-the-art models for summarization are generally neural and transformer-based such as PEGASUS \cite{zhang2020pegasus} and Prophet \cite{qi2020prophetnet}. These models have been used for text summarization for different domains, including news \cite{desai2020compressive} and scientific texts.  For scientific text summarization, \citet{zhang2020pegasus} have obtained best results in existing literature, based on evaluation on a dataset of articles published on arXiv and PubMed using papers' abstracts as ground truth.

Scientific texts pose specific problems for summarization, given their particular structure and way of organizing information. This is why the problem of scientific text summarization has been approached separately from general summarization systems.
The problem of scientific text summarization has been approached before \cite{yasunaga2019scisummnet,altmami2020automatic,ju2020scisummpip,Cohan2017ContextualizingCF,qazvinian2010citation}. Top conferences in NLP have organized workshops on scholarly document processing, including shared tasks specifically focused on scientific document summarization \cite{chandrasekaran2019overview}. Most approaches for scientific text summarization use an extractive \cite{saggion2000selective,saggion2011learning,yang2016amplifying,slamet2018automated,agrawal2019scalable,hoang2010towards} or citation-based approach \cite{Cohan2017ContextualizingCF,qazvinian2010citation,ronzano2016empirical}, with a few exceptions attempting abstractive summarization on scientific texts \cite{lloret2013compendium}. Notably, \citet{ju2020scisummpip} use a combined extractive and abstractive approach based on BERT. \citet{sun2018summarization} propose an approach based on semantic link networks for summarizing scientific texts. A recently published survey \cite{altmami2020automatic} contains a more exhaustive overview of previous attempts at summarizing scientific papers.

Given the excellent results of recent text generation models, it is promising to consider new applications in fields where they have not been leveraged in practice before.
We propose that one such task is {\em scientific review summary generation}. We evaluate in this paper the feasibility of automatically generating review summaries for scientific papers. We use state-of-the-art models for text summarization, and apply them to our problem. We release a dataset of articles and reviews from NeurIPS, which we use to assess the performance of automatic summarization models for the problem of review summary generation.

\section{Dataset}

We build a dataset of articles and associated reviews by scraping NeurIPS's conference website,\footnote{https://papers.neurips.cc} and collecting all articles published in NeurIPS between 2013 and 2020, along with their reviews. To obtain the full text of the papers, we downloaded the PDFs from the website and extracted the text using Grobid.\footnote{https://github.com/kermitt2/grobid} Reviews were extracted directly from the HTML content of the web pages, and, where needed, heuristics were used in order to exclude the texts of the author's responses.  Each article can have several reviews. Table \ref{tab:dataset} summarizes statistics about the dataset.

\begin{table}[]
    \centering
    \small
    \begin{tabular}{r|l}
        
        \textbf{Articles} & 5,950 \\
        \textbf{Reviews} & 18,926 \\
        \textbf{Avg review len (words)} & 399 \\
        \textbf{Avg review len (sentences)} & 21 \\
        \textbf{Avg abstract len (words)} & 159 \\
        \textbf{Avg abstract len (sentences)} & 7 \\
    \end{tabular}
    \caption{Dataset statistics.}
    \vspace{-3mm}
    \label{tab:dataset}
\end{table}


\section{Summarization Experiments}

Reviews of scientific articles are usually comprised of a short summary, followed by the comments comprising the reviewer's evaluation of the article, mentioning its strengths and its weaknesses. The initial summary of the paper is usually a short objective description of its contents, so in theory it could be inferred solely based on the article's content. Based on this premise, we formulate the problem of automatic review generation as a text summarization problem.

\begin{table*}[ht]
    \centering
    \small
    \begin{tabular}{r|c|c|c|c}
        & \textbf{R-1} & \textbf{R-2} & \textbf{R-L} & \textbf{BERTScore} \\ 
        \hline
        \hline 
        \textbf{vs. arXiv abstracts \cite{zhang2020pegasus}} & .447 & .173 & .258 & - \\
        \hline
        \textbf{vs. abstract (NeurIPS)} & .236 & .046 & .151 & .793  \\
        
        \hline
        \textbf{vs. review summaries (individual whole)} & .169 & .023 & .117 &  .789  \\
        \textbf{vs. review summaries (concatenated whole)} & .206 & .033 & .127 & .784

    \end{tabular}
    \caption{Performance of pretrained model}
    \label{tab:eval_pretrain}
\end{table*}

\textbf{Pre-processing.} We aim to separate the two different parts of each review: the initial part containing a short summary of the paper, from the following comments and evaluation of the paper. A manual inspection of extracted reviews in our dataset for papers up to 2019 shows that many reviews include replies to author responses from the rebuttal phase of the review, and these can be found either in the beginning or end of the review, without a consistent pattern, sometimes separated from the main review by ASCII separators (strings of "-"/"="/"~"). We then rely on heuristics in order to correctly extract the summary part of the review, by searching the review text for keywords such as "rebuttal" or "response": if these are found at the beginning of the review, we then look for ASCII separator characters, and consider the original review to begin after the separator; otherwise, we assume the summary is found at the beginning of the review. For papers from NeurIPS 2020, the different sections of the review are clearly marked (\textit{summary}, \textit{strengths}, \textit{weaknesses}, \textit{clarity} and \textit{correctness}), so this pre-processing step was not needed. After this step, we split the obtained text into sentences and select the first $k$ sentences as the summary. Our motivation in doing so was driven by several works on extreme classification \cite{DBLP:journals/corr/abs-1907-08722,narayan-etal-2018-dont} aimed at generating short, one-sentence news summary to answer the question: {``}What is the article about?{''}.


\begin{table*}[ht]
    \centering
    \small
    \begin{tabular}{r|c|c|c|c}
        & \textbf{R-1} & \textbf{R-2} & \textbf{R-L} & \textbf{BERTScore} \\ 
        \hline
        \hline 
        \textbf{vs. abstract (NeurIPS)} & .261 & .034 & .141 & .812 \\
        \hline
        \textbf{vs. review summaries (indiviual whole)} & .230 & .031 & .148 & .817\\
        
        \textbf{vs. review summaries (concatenated whole)} & .254 & .046 & .145 & .806  \\
        \hline
        \textbf{vs. review summaries (concatenated 5 sents)} & .273 & .047 & .155 & .808 \\
        \textbf{vs. review summaries (concatenated 4 sents)} & .279 & .046 & .158 & .810 \\
        \textbf{vs. review summaries (concatenated 3 sents)} & .287 & .045 & .164 & .813\\
        \textbf{vs. review summaries (concatenated 2 sents)} & .290 & .042 & .170 & .817 \\
        \textbf{vs. review summaries (concatenated 1 sent)} & .246 & .032 & .160 & .821\\
        \hline
        \textbf{vs. review summaries (individual 5 sents)} & .227 & .030 & .149 & .818 \\
        \textbf{vs. review summaries (individual 4 sents)} & .220 & .028 & .147 & .819 \\
        \textbf{vs. review summaries (individual 3 sents)} & .207 & .026 & .117 & .819 \\
        \textbf{vs. review summaries (individual 2 sents)} & .176 & .022 & .127 & .820 \\
        \textbf{vs. review summaries (individual 1 sent)} & ,114 & .053 & .091 & .822 \\

    \end{tabular}
    \caption{Performance of fine-tuned model on abstract and review summary}
    \label{tab:eval_finetune}
\end{table*}

\textbf{Model.} Language modeling in NLP has recently seen great advancements, and is one of the most active areas of research in NLP, with new results being published every few months. The best performing models are based on neural architectures, among which transformers play an important role. Text summarization in particular is a type of text generation problem; the current state of the art in text generation is PEGASUS \cite{zhang2020pegasus}, which is a transformers-based model trained to generate summaries by masking important sentences in a source text. PEGASUS obtained state-of-the-art results in text summarization across 12 different datasets in different domains, including scientific texts.

We experiment with using PEGASUS in order to generate summaries of scientific articles in our dataset, and assess its performance compared to the collected reviews. 

\textbf{Model pre-trained on abstracts.} We first experiment with a pre-trained version of PEGASUS for scientific text summarization, which was trained to generate abstracts of scientific texts based on a dataset of arXiv articles \cite{cohan2018discourse}. 
In order to ensure no overlap between the test set used for evaluation in our experiments and the articles in the arXiv database used in pre-training of the model, we select as our test set only the articles in our dataset published in 2020 (the arXiv dataset was published in 2018) - we use 1000 of these articles as our test set and keep the rest of 898 as a validation set. The 2020 reviews are also the highest-quality of our dataset, since the summary section of the review is clearly marked and used as is for evaluation (as opposed to extracted based on heuristics).

\textbf{Model fine-tuned on reviews.} Second, we attempt to generate paper summaries which best approximate a review. For this purpose, we fine-tune the pre-trained model used in the previous experiment on our own data, using as targets the reviews in our dataset. 
As a training set, we use the articles and reviews in our dataset published before 2020. While our dataset is smaller than the arXiv dataset used for the pre-trained model, it is expected to be similar to the original training data. For each article, one review is selected at random and used as ground truth for training the summarization model. The training set contains 4,052 papers and their reviews.

\textbf{Evaluation.} We evaluate the models using the ROUGE metric, and compare the generated summaries both to the abstract and the reviews. We report ROUGE-1, ROUGE-2 and ROUGE-L, as well as BERTScore, using the RoBERTa-large model\footnote{\tt roberta-large\_L17\_no-idf\_version=0.3.9 (hug\_trans=4.2.2)} \cite{zhang2019bertscore}. Our setup 
can be evaluated on multiple labels for the same input text: in our test set, one paper can have several reviews. We  evaluate our models with multiple labels: first by considering them separately as independent examples, and second by concatenating all reviews for a given input article into one single reference text, and evaluating against it.

We show examples of generated reviews using our  model, along with the original reviews for the same article, in the Appendix. 

\textbf{Results.} We report separately the results of the pre-trained and the fine-tuned model. We compare different setups, using as target texts both the abstracts and the reviews. In the case of the reviews, we consider separately as a target test the whole review or only the summary section, varying the number of extracted sentences from 1 to 5, and experiment with the two evaluation setups: concatenating the different reviews corresponding to one article, or considering them as separate test examples.


Tables \ref{tab:eval_pretrain} and \ref{tab:eval_finetune} and show the results for all setups.
The pre-trained model obtains better results when evaluated against abstracts than against reviews, across configurations and metrics. 
Although the pre-trained model was trained to generate abstracts, the fine-tuned model still obtains slightly better results compared to abstracts, suggesting it might solve a relevant domain adaptation aspect.
The fine-tuned model also shows improved results for review summary generation.
In terms of ROUGE scores, the optimal number of sentences of the summary extracted from the review summary seems to be 2 in the concatenated setup, while in the individual setup, the performance increases with the number of sentences. BERTScore strictly decreases with the number of sentences for both setups. Especially in the concatenated setup, using the first 1-2 sentences in the review summary as labels out-performs evaluating against the full review summary, suggesting that the generated summaries generally contain information present in the beginning of the review.
\begin{table}[bt]
    \centering
    \small
    \resizebox{.45\textwidth}{!}{%
    \begin{tabular}{r|c|c|c|c}
        & \textbf{R-1} & \textbf{R-2} & \textbf{R-L} & \textbf{BERT} \\ 
       & & & & \textbf{Score} \\
        \hline
        \hline 
        
        \textbf{vs. full review (concat)} & .152 & .036 & .092 & .803  \\
        \textbf{vs. full review (individual)} &  .241 & .040 & .139 & .806 \\
        \hline
        \textbf{vs. strenghts (concat)} & .270 & .039 & .159 & .815 \\
        \textbf{vs. strengths (individual) }& .200 & .038 & .135 & .820 \\
        \textbf{vs. weaknesses (concat)} & .232 & .028 & .134 & .803 \\
       \textbf{ vs. weaknesses (individual) }& .212 & .027 & .134 & .808 \\

    \end{tabular}
    }
    \caption{Performance of fine-tuned model on full review and other review sections}
    \label{tab:eval_finetune_full}
    
\end{table}
\subsection{Feasibility of Generating Full Reviews}

The fine-tuned model is better at generating review summaries than the pre-trained model, across setups. 

The generation of a full review, including critical interpretations from the reviewers, is a much more challenging problem than generating paper summaries. In order to assess how well a summarization model can approximate a full review, including not only the summary, but also the critical comments sections, we separately evaluate our model using the full reviews as targets, as well as against the separate sections (we consider the \textit{Strengths} and \textit{Weaknesses} sections), as show in Table \ref{tab:eval_finetune_full}. We notice that the performance is generally lower than for the review summary, but still comparable. The \textit{Strengths} section seems to have the most in common with the review summary according the better results.

\section{Conclusions}
 
We have formulated the problem of scientific text review generation, as a novel task in NLP with practical applications for the scientific community. Review generation is related to the text summarization task, but has its own specific features, which is what makes it a difficult problem to solve. We have taken the first steps towards building an automatic system for review generation; and have collected and are releasing a dataset of scientific articles and reviews which can be used for future experimentation into the topic.

We conclude that scientific review generation is a difficult problem, with current performance considerably below that of state-of-the-art text generation models on scientific abstracts. Nevertheless, the small improvements in performance we obtain through fine-tuning the model suggest that the problem might be approachable, and encourage us to continue to study it. We propose that more training data could be useful to obtain better results, as would a more accurate extraction of the summary section of the review.
In the future, we would like to explore a more complex training strategy in order to improve performance, such as multi-task learning (to jointly train the model to generate reviews and abstracts), or conditional text generation, in order to constrain the model to generate review-like texts, while keeping the content relevant to the source article. 


\section{Ethical Considerations}

Our dataset poses no privacy issues.
With regards to the task of paper review generation, it is unclear if generating reviews entirely automatically is desirable from a practical as well as ethical perspective. Instead, we approach the problem summary generation for reviews, in view of a possible computer-assisted process for review generation, which would not exclude humans. We think a computational tool for assisting with the ever-growing burden of reviewing can help the community and eventually lead to higher quality reviews, and hope our paper can encourage discussion on the topic. We leave open to discussion the question of how such a tool could best be integrated in the current review system.

\bibliographystyle{acl-ijcnlp2021-templates/acl_natbib}
\bibliography{summarization}

\clearpage

\appendix


\begin{minipage}{2\linewidth}
\section{Appendix}

\begin{tabularx}{\textwidth}{|X|X|}    
    \textbf{Generated review} & \textbf{Original review}\\
    \hline
         
   thanks to the authors for a very interesting paper. the main contribution that the authors have been able to solve this problem is the transient spike packet model, which can take into account discrete populations of neurons in a very long time invariant networks, and as the authors argue allows for a much better signal transmission and/or slower responses to fluctuations in the brain at all times. this implies that the authors argue allows for a much better model of nontriviality invariant networks, independent of population sizes and/or slower responses to fluctuations in the brain at all times. (...)
         & This paper uses the Drosophila\u2019s olfactory system (ORN to PN to LHN) as an example to demonstrate how the heterogeneity in the intrinsic parameters of single neurons could benefit for information transmission in a feedforward network, which is the main conceptual contribution of this work. Then they extend the idea of heterogeneous neurons to multiple layers of feedforward networks. (...)
    \end{tabularx}
    \captionof{table}{Example generated review, and actual review, for article titled "Lamina-specific neuronal properties promote robust, stable signal propagation in feedforward networks".}
\end{minipage}

\bigskip

\begin{minipage}{2\linewidth}
\begin{tabularx}{\textwidth}{|X|X|}
\textbf{Generated summary} & \textbf{Original abstract} \\
\hline
methods for machine- learning models play an important role in developing, and using predictive models as information on what features were important for a given output for any machine- learning model, enabling us to better understand, validate, and interpret model decisions. $<n>$ researchers have developed a number of attribution methods that provide estimates of importance of input features towards model output for specific types of models, and accurate feature estimates for any machine- learning model. $<n>$ we present a new approach to estimating feature importance for any machine- learning model using causal explanation. $<n>$ this approach can be applied to any machine- learning model since it does not require retraining or adapting the original model to demonstrate importance scores are strongly causal, provided another machine- learning objective to train a model uses test data, and another machine- learning objective to learn accurate output. $<n>$ experiments show that the uncertainty estimates provided by this methodology are significantly more accurate than existing model-agnostic methods, and that the uncertainty scores provided by this methodology are significantly more accurate than existing model-agnostic methods. $<n>$ this approach can be applied to any machine- learning model since it does not require retraining or adapting the original model to demonstrate importance scores are strongly causal

&

Feature importance estimates that inform users about the degree to which given inputs influence the output of a predictive model are crucial for understanding, validating, and interpreting machine-learning models. However, providing fast and accurate estimates of feature importance for high-dimensional data, and quantifying the uncertainty of such estimates remain open challenges. Here, we frame the task of providing explanations for the decisions of machine-learning models as a causal learning task, and train causal explanation (CXPlain) models that learn to estimate to what degree certain inputs cause outputs in another machine-learning model. CXPlain can, once trained, be used to explain the target model in little time, and enables the quantification of the uncertainty associated with its feature importance estimates via bootstrap ensembling. We present experiments that demonstrate that CXPlain is significantly more accurate and faster than existing model-agnostic methods for estimating feature importance. In addition, we confirm that the uncertainty estimates provided by CXPlain ensembles are strongly correlated with their ability to accurately estimate feature importance on held-out data. \\
\end{tabularx}
\onecolumn
\captionof{table}{Example of generated summary and original abstract, using pretrained model (paper title "CXPlain: Causal Explanations for Model Interpretation under Uncertainty")}
\end{minipage}

\begin{table*}
\begin{tabularx}{\textwidth}{|X|X|}
\textbf{Generated summary} & \textbf{Original review} \\
\hline
neural networks are extensively used today in safety critical control systems such as autonomous vehicles and airborne collision avoidance systems. $<n>$ rigorous methods to ensure correct functioning of neural network controlled systems is imperative. $<n>$ verification refers to a broad class of techniques that provide strong guarantees of correctness by exhibiting a proof of abstraction. $<n>$ verification of neural networks has attracted lot of attention in recent years. $<n>$ verification of neural networks has attracted a lot of attention in recent years. $<n>$ this paper focuses on neural networks with rectified linear unit as range of activation function and we investigate the output range problem for feed-forward neural networks with rectified linear unit as range of activation function. $<n>$ broad objective is to investigate techniques to verify neural network controlled physical systems such as autonomous vehicles. $<n>$ verification refers to a broad class of techniques that provide strong guarantees of correctness by exhibiting a proof of abstraction. $<n>$ important verification problem is that safety, wherein one seeks to ensure that the neural network controlled system never reaches an unsafe set of states. $<n>$ important computation is to compute the output of network controller given a set of input valuations. $<n>$ we focus on neural networks with rectified linear unit as range of activation function and we investigate the output range
&

 First of all, my knowledge of formal verification of neural networks is very limited, and I apologize for the limitations this poses on my review.  That said, I found this paper very interesting, well written, and from my limited understanding of the literature, this seems like a novel and highly useful tool in the toolbox for verifying neural network models. I am strongly in favor of acceptance.  My main questions are the following: * It is not clear to me what increase in false positives does the method introduce by relaxing the estimate of the output of the network to a superset. * I would like to see a more formal definition of the algorithm with the \"moving pieces\" (e.g. partitioning strategies) stated more explicitly. Then I would like to have a discussion of the considerations that go into defining these \"moving pieces\". * What are the practical limitations of the method on real-world network sizes and architectures. \\
 \hline
 & 
 Overall, I like the approach in the paper and the theoretical background looks solid (though I didn't check the proofs). My main problem with a paper is in the  experimental part: * just one experiment is considered on rather toy neural network * no comparison with other methods is made  Thus, it is impossible to evaluate the usefulness of the proposed method. \\
 \hline
 & Pros: 1. Paper is well written and easily readable. 2. It presents a novel approach for state space reduction of neural network that could be extended to similar problems.  3. It address an important issue of computational complexity in output range analysis for neural network.   Cons: 1. This paper should explore different partitioning schemes and provide a timing comparison among them. 2. It should provide a metric to compare degree of over approximation as compared with approach in [4] \\

\end{tabularx}
\caption{Example of generated summary and original reviews, using pre-trained model (paper title "Abstraction based Output Range Analysis for Neural Networks")}
\end{table*}

\end{document}